\documentclass[letterpaper, 10pt, conference]{ieeeconf}      

\IEEEoverridecommandlockouts 
\usepackage{cite}
\usepackage{tabularx}
\usepackage{multirow}
\usepackage{amsmath,amssymb,amsfonts}
\usepackage{graphicx}
\usepackage{textcomp}

\DeclareMathOperator{\st}{s.t.}
\usepackage[table,xcdraw]{xcolor}
\usepackage[linesnumbered,ruled,lined]{algorithm2e}
\usepackage{booktabs}
\usepackage{xcolor}
\usepackage{authblk}
\let\labelindent\relax
\usepackage{enumitem}
\usepackage{hyperref}
\usepackage{breqn}
\usepackage{algpseudocode}
\usepackage{caption}
\usepackage{subcaption}
\usepackage{tcolorbox}
\usepackage{array}
\usepackage{makecell}

\usepackage[english]{babel}
\usepackage{amsthm}

\definecolor{my_green}{RGB}{144, 238, 144}

\allowdisplaybreaks

\title{ Trajectory Optimization Under Stochastic Dynamics Leveraging Maximum Mean Discrepancy }
\author{Basant Sharma, Arun Kumar Singh \thanks{. Basant and Arun are with the University of Tartu. This research was in part supported by grant PSG753 from Estonian Research Council, collaboration project LLTAT21278 with Bolt Technologies and project TEM-TA101 funded by European Union and Estonian Research Council.
Emails: aks1812@gmail.com, basantsharma1990@gmail.com. Code: \url{https://github.com/Basant1861/MPC-MMD}} 
}


\begin{document} 

\maketitle
\thispagestyle{empty}
\pagestyle{empty}



\begin{abstract}

This paper addresses sampling-based trajectory optimization for risk-aware navigation under stochastic dynamics. Typically such approaches operate by computing $\tilde{N}$ perturbed rollouts around the nominal dynamics to estimate the collision risk associated with a sequence of control commands. We consider a setting where it is expensive to estimate risk using perturbed rollouts, for example, due to expensive collision-checks. We put forward two key contributions. First, we develop an algorithm that distills the statistical information from a larger set of rollouts to a \textit{reduced-set} with sample size $N<<\tilde{N}$. Consequently, we estimate collision risk using just $N$ rollouts instead of $\Tilde{N}$. Second, we formulate a novel surrogate for the collision risk that can leverage the distilled statistical information contained in the \textit{reduced-set}. We formalize both algorithmic contributions using distribution embedding in Reproducing Kernel Hilbert Space (RKHS) and Maximum Mean Discrepancy (MMD). We perform extensive benchmarking to demonstrate that our MMD-based approach leads to safer trajectories at low sample regime than existing baselines using Conditional Value-at Risk (CVaR) based collision risk estimate. 


\end{abstract}


\section{Introduction}

Risk-aware trajectory optimization provides a rigorous template for assuring safety under stochastic dynamics model \cite{yin2023risk}, \cite{pmlr-v144-wang21b}. There are two key challenges in this regard. First, characterizing the state transition distribution for non-linear systems under arbitrary noise model is often intractable. This in turn, also prevents obtaining analytical optimizer friendly expression for the underlying safety (collision, lane violation) risk. Some existing works linearize the non-linear dynamics and adopt Gaussian noise model to by-pass this intractability \cite{zhu2019chance}, \cite{tahmasebi2024condition}. However, such approximations can lead to incorrect state transition distribution and consequently poor collision risk estimate. Thus, in this paper, we adopt the premise of sampling-based optimization that relies on simulating the stochastic dynamics \cite{yin2023risk}, \cite{chen2023risk}. These class of approaches can be applied to arbitrary noise model and  even non-differentiable black-box constraint functions. As a result, they allow us to relax the linear dynamics and Gaussian noise assumption prevalent in many existing works, e.g, \cite{zhu2019chance}, \cite{farina2016stochastic}.

\begin{figure}[t!]
    \centering
    \includegraphics[scale=0.5]{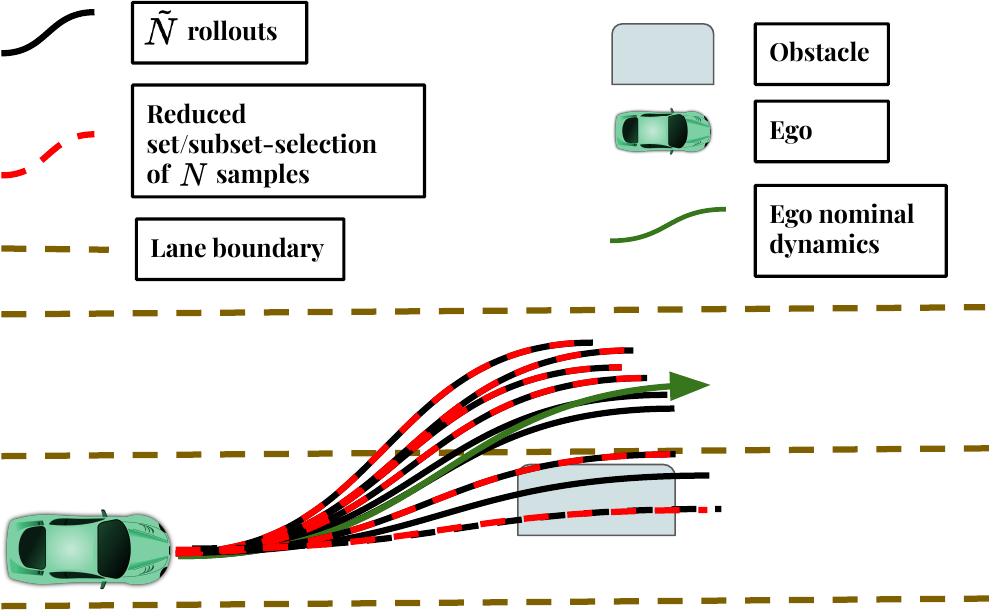}
    \caption{\small{A standard pipeline for risk-aware optimization based on control sampling along with our improvement. These class of approaches rely on simulating the forward dynamics of the vehicle to obtain $\tilde{N}$ samples from the state trajectory distribution, which are then used to estimate risk. Our work provides a novel risk-surrogate and a systematic way of estimating it using a reduced number ($N$) of state trajectory samples (a.k.a the \textit{reduced-set}).   }}
    \label{fig:teaser}
    \vspace{-0.5cm}
\end{figure}

A typical pipeline is presented in Fig.\ref{fig:teaser}. Given a sequence of control commands, we compute $\tilde{N}$ forward simulations a.k.a rollouts of the stochastic dynamics, resulting in as many samples of state trajectories. We then evaluate the state-dependent cost/constraints (e.g safety distance violations) along the state trajectory samples. This is followed by computing some risk-aware statistics such as Conditional Value at Risk (CVaR) of the cost/constraint samples.

In this paper, we consider a setting, where cost/constraint evaluations along the state trajectory samples are computationally expensive, e.g due to expensive collision checks. Furthermore, often, computing the rollouts itself could be computationally and memory intensive. Our key idea is to distill the statistical information contained in $\tilde{N}$ state trajectory samples into a \textit{reduced-set} with sample size $N<< \Tilde{N}$. For example, in our implementation, $\Tilde{N} \approx N^2$. The costly constraint evaluations are then performed only on the reduced-set samples. The main challenge is that the distillation to \textit{reduced-set} should be done in a way that it ensures reliable downstream risk estimation using only the $N$ samples. 

\noindent \textbf{Algorithmic Contribution:} We use distribution embedding in Reproducing Kernel Hilbert Space (RKHS) and the Maximum Mean Discrepancy (MMD) measure \cite{simon2016consistent},\cite{10.5555/2188385.2188410} to formalize our key ideas. Specifically, we develop an MMD-based surrogate for state-dependent risk, leveraging RKHS tools to enhance sample efficiency. Our approach involves a bi-level optimization that systematically reduces the number of state trajectory samples required to estimate the risk of a sequence of control commands. Additionally, we introduce a custom sampling-based optimizer to minimize the proposed risk surrogate.


\noindent \textbf{State-of-the-Art Performance:} We conduct two benchmarking sets to demonstrate the advantages of our approach. First, we show that our MMD-based surrogate is more sample-efficient in estimating collision risk in a given scene compared to popular CVaR-based alternatives \cite{yin2023risk},\cite{lew2023risk},\cite{9802647},\cite{9447796},\cite{9683744}, especially in the low sample regime. Second, we apply our trajectory optimizer in a Model Predictive Control (MPC) setting within the high-fidelity CARLA simulator \cite{Dosovitskiy17}, highlighting improvements over both a deterministic noise-ignorant baseline and a CVaR-based baseline.


\section{Problem Formulation}

\subsubsection*{Symbols and Notations} We use small/upper case normal-font to represent scalars. The bold-face small fonts represent vectors while upper-case variants represent matrices. We use $p_{(.)}$ to denote probability density of $(.)$ and $P$ to represent probability. We use $\langle.,. \rangle_{\mathcal{H}}$ to represent the inner prodct in RKHS $\mathcal{H}$.
\vspace{-0.08cm}
\subsection{Trajectory Optimization}
\noindent We formulate trajectory planning of the ego-vehicle in the road-aligned Frenet frame, described with the help of a reference path (or road center-line). The variable $s$ and $d$ will represent the longitudinal and lateral displacement in the Frenet Frame, while $\psi$ will represent the heading of the vehicle with respect to the reference path. With these notations in place, we define the risk-aware stochastic trajectory optimization in the following manner.
\begin{subequations}
\begin{align}
    \min_{\mathbf{a}, \boldsymbol{\theta}} w_1E[c(\mathbf{x} ) ]+ w_2r(\mathbf{x})+ w_3\left\Vert \begin{matrix}
        \mathbf{a}\\
        \boldsymbol{\theta}
    \end{matrix}\right\Vert_2^2 \label{cost},\\
    \mathbf{x}_{k+1} = f(\mathbf{x}_k, a_{k}+\epsilon_{a, k}, \theta_k+\epsilon_{\theta, k}), \qquad \mathbf{x}_0 \sim p_{0} \label{dynamics_con},\\
     \theta_{min}\leq \theta_k\leq \theta_{max}, a_{min}\leq a_k\leq a_{max} \forall k\label{control_con},
\end{align}
\end{subequations}

\noindent where $c(.)$ represents the state-dependent cost function. The vector $\mathbf{x}_k = (s_k, d_k, \psi_k, \dot{\psi}_k, v_k)$ represents the state of the vehicle at time-step $k$. The vector $\mathbf{x}$ is the concatenation of the states at different $k$. The function $f$ is taken from \cite{liniger2020safe}. The variable $v_k$ is the longitudinal velocity of the ego-vehicle. The control inputs are the longitudinal acceleration $a_k$ and steering input $\theta_k$. The vectors $\mathbf{a}$ and $\boldsymbol{\theta}$ are formed by stacking $a_k$ and $\theta_k$ at different time-step $k$ respectively. The stochastic disturbances acting on the vehicle are modeled as an effect of the perturbation of the nominal acceleration and steering inputs by $\epsilon_{a, k} $ and $\epsilon_{\theta, k} $ respectively. Let $\boldsymbol{\epsilon}_a$, $\boldsymbol{\epsilon}_{\theta}$ be vectors formed by stacking  $\epsilon_{a, k} $ and $\epsilon_{\theta, k} $ respectively at different $k$. We assume that $\boldsymbol{\epsilon}_a\sim p_a$, $\boldsymbol{\epsilon}_{\theta} \sim p_{\theta}$. The perturbation at the control level is mapped to the state trajectory distribution $p_{\mathbf{x}}$ through $f$.  For the sake of generality, we assume that $p_a$ and $p_{\theta}$ are dependent on the control inputs. For example, the slippage of a vehicle on icy-roads depends on the magnitude of the acceleration and steering commands. We don't make any assumptions on the parametric form of $p_{a}, p_{\theta}$ and $p_{\mathbf{x}}$. Instead, we just rely on the ability to sample from $p_{a}, p_{\theta}$  and rollout the dynamics for every sampled ${\epsilon}_{a, k}, {\epsilon}_{\theta, k} $.

The first term in \eqref{cost} minimizes the expected state cost, typically addressing path-following errors where average performance suffices. The second term captures risk in the state trajectory $\mathbf{x}$, considering higher-order noise characteristics like variance, skewness, and kurtosis. Intuitively, it represents the probability of an event (e.g., collision) for a given control sequence. The last term in \eqref{cost} penalizes large control inputs, with weights $w_i$ tuning the ego-vehicle's risk-seeking behavior. Control bounds are enforced through \eqref{control_con}.

\subsection{Algebraic Form of Risk}
\noindent Let $h_k(\mathbf{x}_k)\leq 0, \forall k$ represent state dependent safety constraints. We can eliminate the temporal dependency by defining the worst-case constraint as $h(\mathbf{x}) = \max_k (h_k)$. Clearly, $h(\mathbf{x})\leq 0$ implies $h_k\leq 0, \forall k$. 

In the stochastic setting where $\mathbf{x}_k$ (or $\mathbf{x}$) is a random variable, constraint satisfaction is more appropriately described in terms of so-called chance constraints. These have the general form of $P(h(\mathbf{x})\geq 0)\leq \varepsilon$, where $P$ represents probability and $\varepsilon$ is some constant. Thus, we define risk as 
\begin{align}
    r(\mathbf{x}) = P(h(\mathbf{x})\geq 0)
    \label{risk_def}
\end{align}

\noindent Intuitively, our risk model captures the probability of safety constraints being violated for a given distribution of disturbances. Thus, minimizing it within \eqref{cost}-\eqref{control_con} will lead to safer trajectories. For arbitrary $p_{a}, p_{\theta}$ and/or highly-non linear dynamics model $f$, constraint function $h$, the analytical form of r.h.s of \eqref{risk_def} is often not known. Thus, existing literature on risk-aware trajectory optimization \cite{yin2023risk},\cite{lew2023risk}, \cite{pagnoncelli2009sample} focuses on developing computationally tractable surrogates.

\begin{figure}[!t]
    \centering
    \includegraphics[scale=0.34]{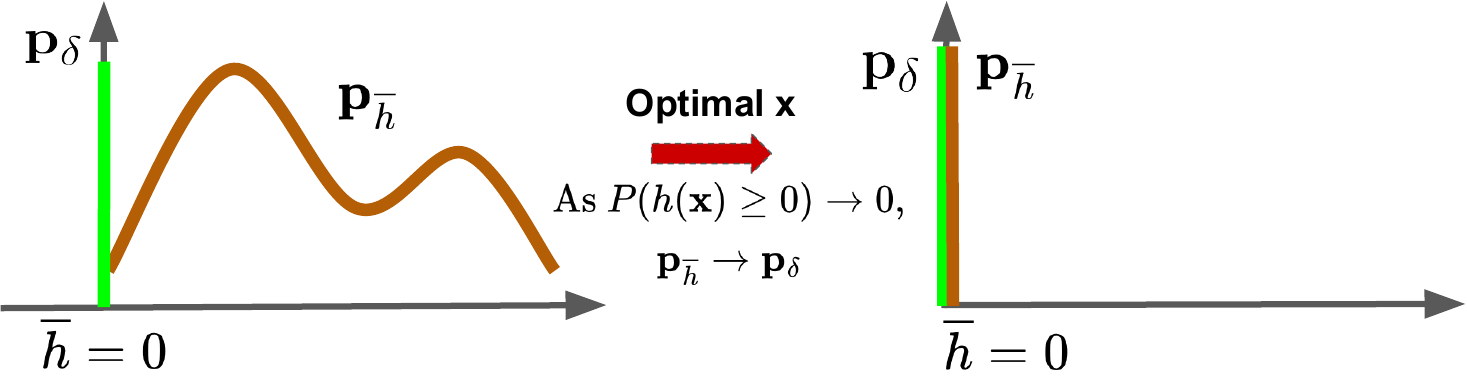}
    \caption{ \footnotesize{The mass of $p_{\overline{h}}$ is to the right of $\overline{h} = 0$}. The optimal control input is one that leads to state-trajectory distribution for which  $p_{\overline{h}}$ resembles a Dirac-Delta distribution.}
    \label{fig:fbar_evolution}
    \vspace{-0.1cm}
\end{figure}

\vspace{-0.1cm}
\section{Main Algorithmic Results}\label{main_results}
Our approach follows the typical pipeline shown in Fig.\ref{fig:teaser}. The unique feature is our risk surrogate and how it can be estimated using only a subset of state-trajectory rollouts. We present the key building blocks next.
\vspace{-0.1cm}
\subsection{State Risk as Difference of Distributions}

\noindent Let us define a constraint residual function as 
\begin{align}
    \overline{h}(\mathbf{x}) = \max(0, h(\mathbf{x})).
    \label{h_bar}
\end{align}

\noindent In the deterministic scenario, driving $\overline{h}(\mathbf{x})$ to zero will push $h(\mathbf{x})$ to the feasible boundary. In the stochastic case, \eqref{h_bar} maps $p_{\mathbf{x}}$ to a distribution of constraint residuals. Let $\overline{h}(\mathbf{x}) \sim p_{\overline{h}}$. The key insight in our work is that although we don't know the parametric form for $p_{\overline{h}}$, we can be certain that its entire mass lies to the right of $\overline{h} = 0$ (see Fig.\ref{fig:fbar_evolution}). Moreover, as $P(h(\mathbf{x})\geq 0)$ approaches zero, the $p_{\overline{h}}$ converges to a Dirac-Delta distribution $p_{\delta}$. In other words, one way of reducing risk is to minimize the difference between $p_{\overline{h}}$ and $p_{\delta}$. Intuitively this minimization will make $p_{\overline{h}}$ look similar to $p_{\delta}$. Thus, we propose the following risk estimate following our prior works \cite{harithas2022cco}, \cite{sharma2023hilbert}.
\begin{align}
    r(\mathbf{x}) \approx \mathcal{L}_{dist}(p_{\delta}, p_{\mathbf{x}}),
    \label{l_dist}
\end{align}
where $\mathcal{L}_{dist}$ is any measure that characterizes the difference between two distributions. For example, Kullback-Leibler Divergence (KLD) quantifies distribution similarity but requires known analytical forms, making it unsuitable for comparing $p_{\overline{h}}$ and $p_\delta$ using only sample-level information. In the following sections, we propose MMD as a potential choice for $\mathcal{L}_{dist}$.
\subsection{RKHS Embedding of Functions of Random Variables}
\noindent Our approach builds on the ability of embedding functions of random variables in the RKHS \cite{simon2016consistent}, \cite{scholkopf2015computing}. For example, state trajectory $\mathbf{x}$ is a function of two random variables $\boldsymbol{\epsilon}_a$, $\boldsymbol{\epsilon}_{\theta}$,  formed by stacking $\epsilon_{a,k}$ and $\epsilon_{\theta, k}$ at different time-step $k$. The RKHS embedding of $\mathbf{x}$ denoted by $\mu[\mathbf{x}]$ is computed as
\begin{align}
    \mu[\mathbf{x}] = E[\phi(\mathbf{x}(\boldsymbol{\epsilon}_{a}, \boldsymbol{\epsilon}_{\theta}))] = E[K_{\sigma}(\mathbf{x}(\boldsymbol{\epsilon}_{a}, \boldsymbol{\epsilon}_{\theta}),.)] ,
    \label{rollout_emb_true}
\end{align}
\noindent where $E[.]$ stands for the expectation operator and $\phi$ is a non-linear transformation commonly referred to as the feature-map \cite{scholkopf2015computing}. One of the key properties of $\phi$ is that the inner product in the RKHS $\left\langle \phi(\mathbf{z}), \phi(\mathbf{z}^')\right\rangle_{\mathcal{H}}$ can be expressed as $K_{\sigma}(\mathbf{z}, \mathbf{z}^')$ for any arbitrary vector $\mathbf{z}, \mathbf{z}^' $. Here,$K_{\sigma}$ is a positive definite function known as the kernel function with hyper-parameter $\sigma$. Throughout this paper, we used the Laplacian kernel for which $\sigma$ represents the kernel-width.

Typically, the r.h.s of \eqref{rollout_emb_true} is difficult to compute. Thus, it is common to compute the empirical estimate by replacing the expectation operator through sample mean in the following manner.
\begin{align}
    \hat{\mu}[\mathbf{x}] = \sum_{i, j = 1}^{i, j = N}\frac{1}{N^2}\phi({^{ij}}\mathbf{x}) = \sum_{i, j = 1}^{i, j = N}\frac{1}{N^2}K_{\sigma}({^{ij}}\mathbf{x}, .),
    \label{empirical_emb}
\end{align}
\noindent where, ${^{ij}}\mathbf{x} = \mathbf{x}({^i}\boldsymbol{\epsilon}_{a}, {^j}\boldsymbol{\epsilon}_{\theta}) $ and
${^i}\boldsymbol{\epsilon}_a$,  ${^j}\boldsymbol{\epsilon}_{\theta}$ are i.i.d samples of $\boldsymbol{\epsilon}_a$, $\boldsymbol{\epsilon}_{\theta}$ respectively. Following a similar approach, the RKHS embedding of  $\overline{h}(\mathbf{x})$ (or $p_{\overline{h}}$) (function of random variable $\mathbf{x}$), and its empirical estimate can be computed as \eqref{hbar_emb}-\eqref{hbar_emb_emp}, wherein, ${^{ij}}\overline{h} = \overline{h}({^{ij}}\mathbf{x})$
\begin{subequations}
\begin{align}
    \mu[\overline{h}] = E[\phi(\overline{h}(\mathbf{x}))], \label{hbar_emb}\\
    \hat{\mu}[\overline{h}] = \sum_{i, j = 1}^{i, j = N}\frac{1}{N^2}\phi ({^{ij}}\overline{h}) =  \sum_{i, j = 1}^{i, j = N}\frac{1}{N^2}K ({^{ij}}\overline{h}, .)  \label{hbar_emb_emp},
\end{align}    
\end{subequations}


\subsection{MMD Based Risk Surrogate}
\noindent Let $\delta$ be a random variable with Dirac-Delta distribution. Let $\mu[\delta]$ be the RKHS embedding of $\delta$ (or $p_{\delta}$). We use the Maximum Mean Discrepancy (MMD) between $p_{\overline{h}}$ and $p_{\delta}$ as our choice for $\mathcal{L}_{dist}$ in \eqref{l_dist} 
\begin{align}
    r_{MMD} = \mathcal{L}_{dist}(p_{\delta}, p_{\mathbf{x}}) = \overbrace{\left\Vert \mu[\overline{h}(\mathbf{x})]-\mu[\delta] \right\Vert_{\mathcal{H}}^2}^{MMD}.
    \label{risk_mmd}
\end{align}

\noindent It can be shown that $r_{MMD} = 0$ implies $p_{\overline{h}} = p_{\delta}$ \cite{simon2016consistent}, \cite{scholkopf2015computing}, \cite{10.5555/2188385.2188410}. In other words, $r_{MMD}=0$ implies that a state trajectory is safe with probability one. From practical stand-point, since we will only have access to state trajectory samples obtained by roll-out of the dynamics, we have to resort to the empirical (biased) MMD estimate \cite{10.5555/2188385.2188410} given by
\begin{align}
    r_{MMD}^{emp} = \left\Vert \hat{\mu}[\overline{h}(\mathbf{x})]-\hat{\mu}[\delta] \right\Vert_{\mathcal{H}}^2,
    \label{risk_mmd_emp}
\end{align}
\noindent where,  $\hat{\mu}[\delta]$ is computed based on the $N^2$ samples of $\delta$ drawn from $p_{\delta}$\footnote{We can approximate $p_{\delta}$ through a Gaussian $\mathcal{N}(0, \epsilon)$, with an extremely small covariance $\epsilon$ ($\approx 10^{-5}$).}. The empirical $r_{MMD}^{emp}$ converges to $r_{MMD}$ at the rate proportional to $\frac{1}{\sqrt{N}}$ \cite{scholkopf2015computing}, \cite{10.5555/2188385.2188410}.

\noindent \textbf{Advantages of $r_{MMD}^{emp}$ as a Risk Estimate:}
\noindent Eqn.\eqref{hbar_emb_emp} suggests that we need $N^2$ rollouts of state trajectory $\mathbf{x}$ and evaluate $\overline{h}(\mathbf{x})$ over each of them to compute $\hat{\mu}[\overline{h}]$ and consequently $r_{MMD}^{emp}$. However, this can be computationally prohibitive if  evaluating $\overline{h}$ is difficult, e.g, due to expensive collision checks. Fortunately, RKHS embedding and MMD provide us a set of tools to systematically choose only $N$ out of those $N^2$ rollouts to estimate $\overline{h}(\mathbf{x})$ and  $r_{MMD}^{emp}$ and yet ensure minimal loss in risk estimation accuracy. 

Let ${^l}\mathbf{x}^', l = 1,2, \dots, N$ be some $N$ subset/\textit{reduced-set} of $N^2$ samples of ${^{ij}}\mathbf{x}$. Moreover, let us re-weight the importance of each ${^l}\mathbf{x}^'$ through ${^l}\beta$ such that $\sum_l {^l}\beta = 1$. Then, the RKHS embeddings of the state trajectory and constraint residual distribution using the ${^l}\mathbf{x}^'$ samples are  given by 
\begin{align}
    \hat{\mu}[\mathbf{x}^'] = \sum_{l = 1}^{l = N}{^l}\beta \phi({^l}\mathbf{x}^'), \qquad \hat{\mu}[\overline{h}^'] = \sum_{l = 1}^{l = N}{^l}\beta \phi(\overline{h}({^l}\mathbf{x}^'))
    \label{x_emd_red}
\end{align}



\noindent Now, Theorem 1 from \cite{simon2016consistent} ensures that if $\hat{\mu}[\mathbf{x}^']$ is close to  $\hat{\mu}[\mathbf{x}]$ in MMD sense, then  $\hat{\mu}[\overline{h}^']$ will be close to $\hat{\mu}[\overline{h}]$ in the same metric. Consequently, $r_{MMD}^{emp}$ computed on the smaller $N$ rollouts will be close to that computed over the larger $N^2$ samples. Alternately, if we can minimize $\left \Vert \hat{\mu}[\mathbf{x}^']-\hat{\mu}[\mathbf{x}]\right\Vert_{\mathcal{H}}^2$ , then we can ensure minimal loss in risk estimation accuracy while reducing the sample-size. It is worth pointing out that such sample optimization feature is not available in risk metrics like CVaR.
\vspace{-0.05cm}
\subsection{Optimal Reduced Set}
\noindent There are three ways in which we can minimize $\left \Vert \hat{\mu}[\mathbf{x}^']-\hat{\mu}[\mathbf{x}]\right\Vert_{\mathcal{H}}^2$. We can choose the optimal subset/\textit{reduced-set} out of $N^2$ dynamics rollouts. Moreover, we can optimize ${^l}\beta$ as well as the Kernel parameter $\sigma$. To this end,  we formulate the following bi-level optimization, wherein $\mathbf{O}$ is a matrix formed by row-wise stacking of the $N^2$ samples of $\mathbf{x}$.
\vspace{-0.05cm}
\begin{subequations}
    \begin{align}\label{bi_level_red_outer}
    \min_{\boldsymbol{\lambda}, \sigma} & \left\Vert  \sum_{i, j = 1}^{i, j = N}\frac{1}{N^2}\phi ({^{ij}}\mathbf{x})-\sum_{l = 1}^{l = {N} }{^l}\beta^* \phi({^l}\mathbf{x}^') \right\Vert_{\mathcal{H}}^2  \\
    & F_{\boldsymbol{\lambda}}({\mathbf{O}}) = ({^1}\mathbf{x}^', {^2}\mathbf{x}^', \dots, {^N}\mathbf{x}^'  ) \\
        & {^l}\beta^*= 
        \begin{aligned}[t]\label{bi_level_inner_red_cost}
        &&& \arg\min_{{^l}\beta} \left\Vert  \sum_{i, j = 1}^{i, j = N}\frac{1}{N^2}\phi ({^{ij}}\mathbf{x})-\sum_{l = 1}^{l = {N} }{^l}\beta \phi({^l}\mathbf{x}^') \right\Vert_{\mathcal{H}}^2 \\
            &&&\st  \sum_l{{^l}\beta} = 1
     \end{aligned}
\end{align}
\end{subequations}

\noindent In the above optimization, ${F}_{\boldsymbol{\lambda}}$ is a function that chooses $N$ rows out of $\mathbf{O}$ to provide the \textit{reduced-set} samples. It is parameterized by vector $\boldsymbol{\lambda}$. That is, different choices of  $\boldsymbol{\lambda}$ lead to different \textit{reduced-set} selection. Note that the cost terms  \eqref{bi_level_red_outer}, \eqref{bi_level_inner_red_cost} can be computed via the kernel trick \cite{simon2016consistent}.

As can be seen, the inner optimization \eqref{bi_level_inner_red_cost} is defined over just the weights ${^l}\beta$ for a fixed \textit{reduced-set} selection given by $\boldsymbol{\lambda}$. The outer optimization in turn optimizes in the space of $\boldsymbol{\lambda}$ and kernel parameter in order to reduce the MMD cost associated with optimal  ${^l}\beta^*$. We use the approach presented in \cite{singh2022bileveloptimizationaugmentedconditional} that combines gradient-free cross entropy method (CEM) \cite{botev2013cross} with quadratic programming (QP) to solve \eqref{bi_level_red_outer}-\eqref{bi_level_inner_red_cost}. Specifically, we sample $\boldsymbol{\lambda}, \sigma$ from a Gaussian distribution and solve the inner optimization for each of the samples to obtain ${^l}\beta^*$. We then evaluate the upper-level cost for all $({^l}\beta^*, \boldsymbol{\lambda}, \sigma)$ and subsequently modify the sampling distribution to push down this cost. We leverage the fact that the inner optimization \eqref{bi_level_inner_red_cost} is essentially a equality constrained QP with a closed-form solution to develop a heavily parallelized solver over GPUs.

\noindent \textbf{Selection Function:} Let $\boldsymbol{\lambda}\in \mathbb{R}^{N^2}$ be an arbitrary vector and $\lambda_t$ be its $t^{th}$ element. Assume that $\vert \lambda_t\vert $ encodes the value of choosing the sample of the state trajectory $\mathbf{x}$ stored in the
$t^{th}$ row of ${\mathbf{O}}$. That is, larger the $\vert \lambda_t\vert $, the higher the impact of choosing the $t^{th}$ row of ${\mathbf{O}}$ in minimizing \eqref{bi_level_red_outer}. With these constructions in place, we can now define $F_{\boldsymbol{\lambda}}$ through the following sequence of mathematical operations.
\begin{subequations}
    \begin{align}
        \mathcal{S} = Argsort\{\vert\lambda_1\vert, \vert\lambda_2\vert, \dots \vert\lambda_{N^2}\vert)\} \label{argsort}\\
        \mathbf{O}^{'} = [ {\mathbf{O}}_{t_1}, {\mathbf{O}}_{t_2}, \dots, {\mathbf{O}}_{t_{N^2}}  ], \forall t_e \in \mathcal{S} \label{sort}  \\
        F_{\boldsymbol{\lambda}}({\mathbf{O}}) = \mathbf{O}^{'}_{t_{N^2-N}: t_{N^2}} =  ({^1}\mathbf{x}^', {^2}\mathbf{x}^', \dots, {^N}\mathbf{x}^'  )\label{selection}
    \end{align}
\end{subequations}

\noindent As can be seen, we first sort in increasing value of $\vert\lambda_t\vert$ and compute the required indices $t_e$. We then use the same indices to shuffle the rows of $\mathbf{O}$ in \eqref{sort} to form an intermediate variable $\mathbf{O}^{'}$. Finally, we choose the last $N$ rows of $\mathbf{O}^{'}$ as our \textit{reduced-set} in \eqref{selection}. Intuitively, \eqref{argsort}-\eqref{selection} parameterizes the sub-selection of the rows of ${\mathbf{O}}$ to form the \textit{reduced-set}. That is, different $\boldsymbol{\lambda}$ gives us different \textit{reduced-sets}. Thus, a large part of solving \eqref{bi_level_red_outer}-\eqref{bi_level_inner_red_cost} boils down to arriving at the right $\boldsymbol{\lambda}$. As mentioned above, this is done through a combination of gradient-free search and quadratic programming.

\subsection{Trajectory Optimizer}
Alg.\ref{algo_1} presents our approach for solving \eqref{cost}-\eqref{control_con} when the risk cost is given by $r_{MMD}^{emp}$. It combines constrained gradient-free Cross Entropy Method (CEM) \cite{singh2022bileveloptimizationaugmentedconditional}, Model Predictive Path Integral (MPPI) \cite{bhardwaj2022storm}, and convex optimization to iteratively refine low-risk, low-cost control inputs.

The algorithm initializes the sampling distribution (Line 2) and samples longitudinal velocity and lateral offset setpoints (Line 5), which are fed to a Frenet planner \cite{wei2014behavioral} (Line 6) to generate trajectories. Using differential flatness \cite{han2023efficient}, these are converted to accelerations and clipped to control bounds. Control perturbations are sampled, and $N^2$ rollouts of \eqref{dynamics_con} yield state trajectory samples (Line 7). A \textit{reduced-set} of $N$ trajectories is selected (Line 8), and $r_{MMD}^{emp}$ is estimated (Line 9). The lowest-risk $n_c$ samples form $ConstraintElliteSet$ (Line 10), from which costs are computed (Line 11) and stored (Line 12). A final $ElliteSet$ of $n_e$ samples is selected (Line 14) to update the sampling distribution (Line 15) via:

\small
\begin{subequations}
\begin{align}
    {^{m+1}}\boldsymbol{\nu} = (1-\eta){^{m}}\boldsymbol{\nu}+\eta\frac{\sum_{q=1}^{q=n_{e}} t_q\mathbf{b}_{q}   }{\sum_{q=1}^{q=n_{e}} t_q}, \label{mean_update}\\
    {^{m+1}}\boldsymbol{\Sigma} = (1-\eta){^{m}}\boldsymbol{\Sigma}+\eta\frac{ \sum_{q=1}^{q=n_{e}} t_q(\mathbf{b}_{q}-{^{m+1}}\boldsymbol{\nu})(\mathbf{b}_{q}-{^{m+1}}\boldsymbol{\nu})^T}   {\sum_{q=1}^{q=n_e} t_q} \label{cov_update}\\
    t_q = \exp{\frac{-1}{\gamma}(c_q   }) \label{s_formula}
\end{align}
\end{subequations}
\normalsize

Here, $\gamma$ is the MPPI temperature parameter \cite{bhardwaj2022storm}, and $\eta$ controls learning rate. By parameterizing long-horizon control sampling using low-dimensional velocity and offset setpoints, Alg.\ref{algo_1} significantly improves computational efficiency.
\noindent 
\begin{algorithm}[!t]
\caption{Sampling-Based Optimizer to Solve \eqref{cost}-\eqref{control_con}}
\small
\label{algo_1}
\SetAlgoLined
$M$ = Maximum number of iterations\\
Initiate mean $^{m}\boldsymbol{\nu}, ^{m}\boldsymbol{\Sigma}$, at iteration $m=0$ for sampling frenet parameters (velocity and lane-offsets) $\mathbf{b}$\\
\For{$m=1, m \leq M, m++$}
{
     \vspace{0.1cm}
    Initialize $CostList$ = []\\
     \vspace{0.1cm}
    
    Draw ${n}$ samples $(\mathbf{b}_{1}, \mathbf{b}_{2}, \mathbf{b}_{q}, ...., \mathbf{b}_{{n}})$ from $\mathcal{N}(^{m}\boldsymbol{\nu}, ^{m}\boldsymbol{\Sigma})$ \\
    
    Query Frenet Planner  $\forall \mathbf{b}_{q}$ : $(\mathbf{a}_q, \boldsymbol{\theta}_q) = \text{Frenet Planner}(\mathbf{b}_q), \forall q = (1, 2, \dots, n)$ \\
    \vspace{0.1cm}
    
    Compute $N$ samples each of $\boldsymbol{\epsilon}_a, \boldsymbol{\epsilon}_{\theta}$ and subsequently $N^2$ rollouts ${^{ij}}\mathbf{x}_q$ for  $(\mathbf{a}_q, \boldsymbol{\theta}_q)$ control trajectory. Repeat this process $\forall q = (1, 2, \dots, n)$ \\  
    
    Choose $N$ rollouts ${^l}\mathbf{x}^'_q$ out of $N^2$ ${^{ij}}\mathbf{x}_q$ through \eqref{bi_level_red_outer}-\eqref{bi_level_inner_red_cost} and compute corresponding ${^l}\beta_q$ and kernel parameter $\sigma_q$. Repeat this process $\forall q = (1, 2, \dots, n)$ \\
    
    Compute $\hat\mu[\overline{h}^']$ over the optimal \textit{reduced-set} through \eqref{x_emd_red} and subsequently $r_{MMD}^{emp}$. Repeat this $\forall q = (1, 2, \dots, n)$ \\
    
    
    $ConstraintEliteSet  \gets$ Select top $n_{c}$ batch of $\mathbf{a}_{q}, \boldsymbol{\theta}_q$, ${^{ij}\mathbf{x}_q}$, $\mathbf{b}_q$ with lowest $r_{MMD}^{emp}$\\
    \vspace{0.1cm}
    
    Define $c_q = w_1\sum_{i, j = 1}^{i, j = N}c({^{ij}}\mathbf{x}_q ) + w_2r_{MMD}^{emp}+w_3\left\Vert \begin{matrix}
            \mathbf{a}_q\\
            \boldsymbol{\theta}_{ q}
        \end{matrix}\right\Vert_2^2 $\\
    $cost \gets$ $c_q$,  $\forall q$ in the $ConstraintEliteSet$ \\
    \vspace{0.1cm}
    
    append each computed ${cost}$ to $CostList$ \\
    \vspace{0.1cm}
        
    $EliteSet  \gets$ Select top $n_{e}$ samples of ($\mathbf{a}_{q}, \boldsymbol{\theta}_{q } $), ${^{ij}\mathbf{x}_q}$, $\mathbf{b}_q$  with lowest cost from $CostList$.\\
     \vspace{0.1cm}
    $({^{m+1}}\boldsymbol{\nu}, {^{m+1}}\boldsymbol{\Sigma} ) \gets$ Update distribution based on $EliteSet$ 
}
\Return{ Control Inputs $\mathbf{a}_{q}$ and  $\boldsymbol{\theta}_{q}$ corresponding to lowest cost in the $EliteSet$}
\normalsize
\end{algorithm}

\vspace{-0.1cm}
\subsection{Diagonal Estimation of RKHS Embedding}
\noindent The RKHS embedding computed in \eqref{empirical_emb} is defined in the product space formed with $N$ samples each of noise $\boldsymbol{\epsilon}_{a}$ and $\boldsymbol{\epsilon}_{\theta}$. For example, if $(^1\boldsymbol{\epsilon}_{a}, {^2}\boldsymbol{\epsilon}_{a})$ and $(^1\boldsymbol{\epsilon}_{\theta}, {^2}\boldsymbol{\epsilon}_{\theta})$ are respectively samples of $\boldsymbol{\epsilon}_{a}$ and $\boldsymbol{\epsilon}_{\theta}$, then we form noise pairs such as $(^{1}\boldsymbol{\epsilon}_{a}, ^{1}\boldsymbol{\epsilon}_{\theta})$, $(^{2}\boldsymbol{\epsilon}_{a}, ^{1}\boldsymbol{\epsilon}_{\theta})$, etc. This in turn results in $N^2$ state trajectory rollouts. However, it is possible to use the so-called diagonal estimation of RKHS embedding that requires only $N$ state trajectory rollouts \cite{simon2016consistent}. It is given by:
\begin{align}
    \hat{\mu}_D[\mathbf{x}] = \sum_{i= 1}^{i= N}\frac{1}{N}\phi({^{i}}\mathbf{x}), \qquad {^{i}}\mathbf{x} = \mathbf{x}({^i}\boldsymbol{\epsilon}_{a}, {^i}\boldsymbol{\epsilon}_{\theta}) 
    \label{empirical_emb_diagonal}
\end{align}
The embedding $\hat{\mu}_D[\mathbf{x}]$ \eqref{empirical_emb_diagonal} has a higher variance than $\hat{\mu}[\mathbf{x}]$  that also translates to the MMD-based risk estimation. We will refer to the risk cost computed using $\hat{\mu}_D[\mathbf{x}]$ as $r_{MMD-D}^{emp}$. Nevertheless, in very low sample regimes, $\hat{\mu}_D[\mathbf{x}]$ can often give competitive performance and we analyze this further in Section \ref{validation}. 

\section{Connections to Related Works}
\noindent \textbf{Linear Dynamics and Constraints Under Gaussian Noise:} For linear dynamics perturbed by Gaussian noise and affine per-step constraint function $h_k$, the risk defined in \eqref{risk_def} has an exact convex reformulation \cite{farina2016stochastic}, allowing for efficient trajectory optimization. For non-linear systems and constraint functions, linearization can achieve a similar structure \cite{zhu2019chance}, \cite{castillo2020real}, but this may lead to inaccuracies in the state and constraint distribution, compromising safety guarantees. Recently, \cite{wang2020eb} introduced tractable reformulations for $r$ under Gaussian Mixture Models. In contrast, our work only need sample level information of the uncertainty and  imposes no assumptions on the algebraic form of the constraints, or dynamics.


\noindent \textbf{Arbitrary Dynamics, Constraints and Noise Model:} In this context, the risk $r$ lacks an analytical form, leading existing works to approximate it using samples drawn from the state distribution to evaluate the constraint function. For instance, \cite{lew2023risk}, \cite{9802647}, \cite{9447796}, and \cite{9683744} estimate $r$ through Conditional Value at Risk (CVaR) over samples of $h({^{ij}}\mathbf{x})$. Similarly, \cite{yin2023risk} defines $r$ as CVaR over the constraint residual function $\overline{h}({^{ij}}\mathbf{x})$. Another approach involves representing $r$ through the sample average approximation (SAA) of 
$\overline{h}({^{ij}}\mathbf{x})$ samples \cite{pagnoncelli2009sample}.

A key differentiating factor between existing CVaR \cite{yin2023risk}, \cite{lew2023risk}, \cite{9802647}, \cite{9447796}, \cite{9683744}, and SAA \cite{pagnoncelli2009sample} approximations and our MMD-based risk surrogate is our ability to maximize the expressive capacity of the latter for a given sample size (see \eqref{bi_level_red_outer}-\eqref{bi_level_inner_red_cost}). We are not aware of any similar mechanism in existing works for CVaR and SAA-based risk approximations.


\noindent \textbf{Chance-Constrained Optimization:} Instead of minimizing risk $r$, we can enforce constraints of the form $r \leq \gamma$ for some $\gamma \in [0,1]$, leading to a more restrictive chance-constrained optimization setting. These problems are typically tractable only for linear dynamics, Gaussian uncertainty, and affine $h(\mathbf{x})$, while more general cases often rely on CVaR-based reformulations \cite{lew2023risk}. As discussed in Section \ref{main_results}, our MMD-based risk serves as a surrogate for $r$, aiming to minimize it rather than enforcing an upper bound, which may be difficult to determine beforehand.

\noindent \textbf{Connections to Works on RKHS Embedding:} Our formulation builds upon the RKHS embedding approach presented in \cite{simon2016consistent}, \cite{scholkopf2015computing} for functions of multiple random variables. These works also suggest using the concept of \textit{reduced-set} to maximize the expressive capacity of the embeddings. However \cite{simon2016consistent}, \cite{scholkopf2015computing} relies on randomly selecting a subset of the samples as the \textit{reduced-set}. In contrast, our bi-level optimization \eqref{bi_level_red_outer}-\eqref{bi_level_inner_red_cost} provides a much more principled approach. Our proposed work also extends the MMD-based risk surrogate presented in \cite{harithas2022cco}, \cite{sharma2023hilbert} to the case of stochastic dynamics. A slightly different perspective from our approach is the setting of Distributionally Robust Optimization (DRO). It is typically employed when the  underlying probability distribution is itself not known accurately and operates by by optimizing over an ambiguity set of possible distributions. In contrast, our work assumes a well-estimated noise distribution for direct risk modeling. However, our method can be extended to DRO using MMD-based ambiguity sets \cite{9993212}, offering a promising avenue for future research.


\section{Validation and Benchmarking}\label{validation}
This section compares trajectory optimization using $r_{MMD}^{emp}$ as the risk cost against other popular alternatives while also highlighting the inner workings of our MMD-based approach.

\vspace{-0.1cm}
\subsection{Implementation Details}
\noindent We implemented optimization \eqref{bi_level_red_outer}-\eqref{bi_level_inner_red_cost} and Alg.\ref{algo_1} in Python using Jax \cite{jax2018github} as the GPU-accelerated linear algebra back-end. Our constraint function $h(\mathbf{x})$ has two parts. For the first part, we check if every lateral position $d_k$ is within the lane bounds. The second part computes the distance of the ego-vehicle with the neighboring vehicles, by modeling each of them as ellipsoids. We handle multiple obstacles by simply taking the worst-case collision distance over all the obstacles \cite{lew2023risk}.  To ensure the reproducibility of our benchmarking, we consider only static obstacles in our comparative analysis and we assume that their positions are known. But in the accompanying video, we show results with dynamic obstacles as well.


The state cost has the following form
\small
\begin{align}
    c(\mathbf{x}) = \sum_k (v_k-v_d)^2+\vert(d_k-d_1)\vert\vert(d_k-d_2)\vert+(\ddot{s}_k^2+\ddot{d}_k^2),
\end{align}
\normalsize

\noindent where $v_d$ is the desired forward velocity and $d_1, d_2$ are two lane center-lines which the ego-vehicle can choose to follow at any given time. The last two terms penalize value in the second-order position derivatives.

\begin{table}[h]
    \centering
    \captionsetup{justification=centering}
    \caption{\footnotesize{(3 Static Obstacles) Low Gaussian noise: $c_{a,1}=0.1, c_{\theta,1}=0.1$. High Gaussian noise: $c_{a,1}=0.15, c_{\theta,1}=0.15$. Low Beta noise: $c_{a,1}=0.1, c_{\theta,1}=0.001$. High Beta noise: $c_{a,1}=0.15, c_{\theta,1}=0.0015$. $c_{a,2}=0.001$, $c_{\theta,2}=0.001$}}
    \vspace{2mm} 
    \renewcommand{\arraystretch}{1.5} 
    \resizebox{\columnwidth}{!}{%
    \begin{tabular}{|>{\centering\arraybackslash}p{2cm}|>{\centering\arraybackslash}p{2cm}|
                    >{\centering\arraybackslash}p{1.1cm}|>{\centering\arraybackslash}p{1.1cm}|
                    >{\centering\arraybackslash}p{1.1cm}|>{\centering\arraybackslash}p{1.1cm}|
                    >{\centering\arraybackslash}p{1.1cm}|>{\centering\arraybackslash}p{1.1cm}|}
        \hline
        \multirow{3}{*}{\textbf{Noise}} & \multirow{3}{*}{\textbf{Method}} & \multicolumn{6}{c|}{\textbf{($\%$ Collisions)}} \\  
        \cline{3-8}
        & & \multicolumn{2}{c|}{\textbf{N=2}} & \multicolumn{2}{c|}{\textbf{N=4}} & \multicolumn{2}{c|}{\textbf{N=6}} \\  
        \cline{3-8}
        & & \textbf{Median} & \textbf{Worst} & \textbf{Median} & \textbf{Worst} & \textbf{Median} & \textbf{Worst} \\  
        \hline
        \multirow{3}{*}{Low Gaussian} & ${r}_{MMD}^{emp}$ & \textbf{1.8} & 23.4 & 0.85 & 9.5 & \textbf{0.2} & \textbf{4} \\  
        \cline{2-8}                     
                                    & $r_{MMD-D}^{emp}$ & \textbf{1.8} & \textbf{22.9} & \textbf{0.7} & \textbf{7.5} & 0.6 & 8.4 \\ 
        \cline{2-8}  
                                    & $r_{CVaR}^{emp}$ & 4 & 32.5 & 1.25 & 16.4 & 1.35 & 10.4 \\ 
        \specialrule{2pt}{0pt}{0pt}
        \multirow{3}{*}{High Gaussian} & $r_{MMD}^{emp}$ & \textbf{3.9} & \textbf{31.4} & 2.8 & 15 & 1 & \textbf{9} \\  
        \cline{2-8}                     
                                    & $r_{MMD-D}^{emp}$ & 4.2 & 32.2 & \textbf{1.8} & \textbf{10.9} & \textbf{0.9} & 9.3 \\ 
        \cline{2-8}  
                                       & $r_{CVaR}^{emp}$ & 5 & 39.9 & 3.1 & 17.7 & 1.7 & 11.5 \\ 
        \specialrule{2pt}{0pt}{0pt}
        \multirow{3}{*}{Low Beta} & $r_{MMD}^{emp}$ & \textbf{0} & 5 & \textbf{0} & 1.7 & \textbf{0} & \textbf{1.5} \\  
        \cline{2-8}                     
                                    & $r_{MMD-D}^{emp}$ & \textbf{0} & \textbf{4.5} & \textbf{0} & \textbf{1.6} & \textbf{0} & \textbf{1.5} \\ 
        \cline{2-8}  
                                   & $r_{CVaR}^{emp}$ & 0.3 & 18.5 & \textbf{0} & 5.4 & 0 & 3 \\ 
        \specialrule{2pt}{0pt}{0pt}
        \multirow{3}{*}{High Beta} & $r_{MMD}^{emp}$ & 0.3 & 10 & \textbf{0} & 3.3 & \textbf{0} & \textbf{1.5} \\  
        \cline{2-8}                     
                                    & $r_{MMD-D}^{emp}$ & \textbf{0.2} & \textbf{9.2} & \textbf{0} & \textbf{2.5} & \textbf{0} & 2.3 \\ 
        \cline{2-8}  
                                    & $r_{CVaR}^{emp}$ & 0.5 & 21.8 & 0.4 & 8.1 & 0.25 & 6 \\ 
        \hline
    \end{tabular}
    }
    \label{tab:synthetic_static}
\end{table}

\begin{table}[h]
    \centering
    \captionsetup{justification=centering}
    \caption{ \footnotesize{(1 dynamic obstacle) Low Gaussian noise: $c_{a,1}=0.1, c_{\theta,1}=0.1$. High Gaussian noise: $c_{a,1}=0.15, c_{\theta,1}=0.15$. Low Beta noise: $c_{a,1}=0.1, c_{\theta,1}=0.005$. High Beta noise: $c_{a,1}=0.15, c_{\theta,1}=0.0075$. $c_{a,2}=0.001$, $c_{\theta,2}=0.001$}}
    \vspace{2mm} 
    \renewcommand{\arraystretch}{1.5} 
    \resizebox{\columnwidth}{!}{%
    \begin{tabular}{|>{\centering\arraybackslash}p{2cm}|>{\centering\arraybackslash}p{2cm}|
                    >{\centering\arraybackslash}p{1.1cm}|>{\centering\arraybackslash}p{1.1cm}|
                    >{\centering\arraybackslash}p{1.1cm}|>{\centering\arraybackslash}p{1.1cm}|
                    >{\centering\arraybackslash}p{1.1cm}|>{\centering\arraybackslash}p{1.1cm}|}
        \hline
        \multirow{3}{*}{\textbf{Noise}} & \multirow{3}{*}{\textbf{Method}} & \multicolumn{6}{c|}{\textbf{($\%$ Collisions)}} \\  
        \cline{3-8}
        & & \multicolumn{2}{c|}{\textbf{N=2}} & \multicolumn{2}{c|}{\textbf{N=4}} & \multicolumn{2}{c|}{\textbf{N=6}} \\  
        \cline{3-8}
        & & \textbf{Median} & \textbf{Worst} & \textbf{Median} & \textbf{Worst} & \textbf{Median} & \textbf{Worst} \\  
        \hline
        \multirow{3}{*}{Low Gaussian} & $r_{MMD}^{emp}$ & \textbf{4.1} & 41.3 & \textbf{0.25} & \textbf{9.2} & \textbf{0.55} & 9 \\  
        \cline{2-8}                     
                                    & $r_{MMD-D}^{emp}$ & 4.4 & \textbf{35.1} & 1.35 & 13.2 & 0.8 & \textbf{8.7} \\ 
        \cline{2-8}  
                                    & $r_{CVaR}^{emp}$ & 6.4 & 43.1 & 2.7 & 20.1 & 1.5 & 12.3 \\ 
        \specialrule{2pt}{0pt}{0pt}
        \multirow{3}{*}{High Gaussian} & $r_{MMD}^{emp}$ & 7.8 & 49.2 & \textbf{1.8} & \textbf{11.1} & \textbf{1.35} & 11.4 \\  
        \cline{2-8}                     
                                    & $r_{MMD-D}^{emp}$ & \textbf{6.7} & \textbf{40.7} & 3 & 17.2 & 1.4 & \textbf{10.4} \\ 
        \cline{2-8}  
                                       & $r_{CVaR}^{emp}$ & 9 & 46.3 & 3.5 & 19.7 & 1.6 & 11.3 \\ 
        \specialrule{2pt}{0pt}{0pt}
        \multirow{3}{*}{Low Beta} & $r_{MMD}^{emp}$ & \textbf{3.95} & \textbf{37.4} & \textbf{0.6} & \textbf{12} & \textbf{0.5} & \textbf{8.9} \\  
        \cline{2-8}                     
                                    & $r_{MMD-D}^{emp}$ & 4.3 & 40.7 & 1.8 & 16.2 & 0.8 & 11.1 \\ 
        \cline{2-8}  
                                   & $r_{CVaR}^{emp}$ & 8.5 & 55.6 & 3.5 & 28.3 & 1.6 & 12 \\ 
        \specialrule{2pt}{0pt}{0pt}
        \multirow{3}{*}{High Beta} & $r_{MMD}^{emp}$ & \textbf{5.1} & \textbf{35.2} & \textbf{0.65} & \textbf{12.9} & \textbf{1.4} & \textbf{15.2} \\  
        \cline{2-8}                     
                                    & $r_{MMD-D}^{emp}$ & 13 & 78.9 & 5.85 & 39.5 & 2.4 & 18.4 \\ 
        \cline{2-8}  
                                    & $r_{CVaR}^{emp}$ & 17.3 & 80.4 & 5.6 & 36.3 & 3.5 & 22.2 \\ 
        \hline
    \end{tabular}
    }
    \label{tab:synthetic_dynamic}
\end{table}

\subsubsection{Benchmarking Environment} We benchmark in trajectory optimization and Model Predictive Control (MPC) settings. For trajectory optimization, we randomly sample initial states for the ego vehicle, static obstacles, and dynamic obstacle trajectories, using Alg.\ref{algo_1} to compute the optimal trajectory. In the MPC setting, we continually re-plan with Alg.\ref{algo_1} based on the ego vehicle's current state.

Using the high-fidelity simulator CARLA\cite{Dosovitskiy17}, we conduct all MPC experiments where the ego vehicle maneuvers from a start to a goal position along a given reference path. An experiment is deemed successful if the vehicle completes the run without collisions. We test in town 10 (T10) and town 05 (T5), with a two-lane scenario featuring 8 obstacles in T10 and 10 in T5. We also introduce uncertainty in the ego vehicle's dynamics through perturbations in nominal acceleration and steering control inputs.



The Gaussian noise distribution has the following form:
\begin{subequations}
\begin{align}
    \epsilon_{a, k} \sim \vert c_{a, 1}a_k\vert \mathcal{N}(0, 1)+c_{a, 2}\mathcal{N}(0, 1), \label{acc_noise_gaussian}\\
    \epsilon_{\theta, k} \sim \vert c_{\theta, 1}\theta_k\vert \mathcal{N}(0, 1)+c_{\theta, 2} \mathcal{N}(0, 1) \label{theta_noise_gaussian},
\end{align}    
\end{subequations}
\noindent where $c_{a, i}$ and $c_{\theta, i}$ are positive constants. As can be seen, the noise have a part that depends on the magnitude of the control and a part that is constant. We create different noise settings by varying  $c_{a, i}$ and $c_{\theta, i}$. We also consider a setting where the r.h.s of \eqref{acc_noise_gaussian}, \eqref{theta_noise_gaussian} are replaced by Beta distribution with probability density function $g(x;a,b)\propto x^{a-1}(1-x)^{b-1}$ where $a,b$ are control dependent parameters. The Beta noise distribution has the following form:
\begin{subequations}
\begin{align}
    \epsilon_{a, k} \sim c_{a, 1} \mathcal{B}(a, b)+c_{a, 2}\mathcal{N}(0, 1), \label{acc_noise_beta}\\
    \epsilon_{\theta, k} \sim c_{\theta, 1} \mathcal{B}(a, b)+c_{\theta, 2} \mathcal{N}(0, 1) \label{theta_noise_beta},\\
    a=\{2\vert a_{k} \vert, 2\vert \theta_k \vert \}, b=\{5\vert a_{k} \vert, 5\vert \theta_k \vert \}
\end{align}    
\end{subequations}
\subsubsection{Baselines}\label{baselines} We consider three baselines. \\
\noindent \textbf{Derterministic (DET):} This is a noise ignorant approach. We use \cite{singh2022bileveloptimizationaugmentedconditional} as the planner as its structure is very similar to Alg.\ref{algo_1}. \\
\noindent \textbf{Ours with $r_{MMD-D}^{emp}$}: This is similar to our main approach, except that the RKHS embedding and the MMD-based risk cost is constructed using the diagonal estimation derived in \eqref{empirical_emb_diagonal}.\\ 
\noindent \textbf{CVaR based Approach: } This baseline replaces $r_{MMD}^{emp}$ with a CVaR based risk in Alg.\ref{algo_1}.
\begin{align}
   r(\mathbf{x}) \approx r_{CVaR}^{emp} = CVaR^{emp}(\overline{h} (\mathbf{x}) )\label{cvar_emp}
\end{align}
\noindent where $CVaR^{emp}$ is the empirical CVaR estimate computed from constraint residual samples $\overline{h}({^{ij}}\mathbf{x})$ along the state-trajectory rollouts \cite{yin2023risk}. Since both $r_{MMD}^{emp}$ and $r_{CVaR}^{emp}$ use the same trajectory optimizer, our benchmarking minimizes optimizer bias to fairly evaluate their finite-sample performance.


\subsubsection{Metrics}\label{metrics} 
We use the following 3 different metrics to validate our results as well as compare against the baseline. 

\noindent \textbf{Collision percentage:} In the trajectory optimization setting, we rollout the vehicle states by perturbing the optimal trajectory computed from Alg.\ref{algo_1}. We perform a large number of rollouts to estimate the ground-truth collision-rate. In the MPC setting, we calculate the number of collisions out of 50 experiments ran in each of town T10 and T05. Each MPC run, consists of around 2000 re-planning steps.\\
\noindent \textbf{Lane constraints violation percentage:}. We don't consider any lane-constraints in the trajectory optimization setting. In the MPC setting, for each experiment we add the lateral lane violations(in metres) at each MPC step and subsequently divide it by the total arc length of the reference trajectory. We then take the average across all experiments. \\
\noindent \textbf{Average and Maximum speed(m/s):} For average speed, we add the ego speed at each MPC step for each experiment and then divide by the total number of steps in that particular experiment. Subsequently we take the average across all experiments. For maximum speed, we compute the highest speed achieved in each MPC experiment and subsequently average it across experiments.
\newtheorem{remark}{Remark}\label{rem_1}
\begin{remark}
    In all our experiments, we fix the number of state-trajectory samples $N$ along which we evaluate $h(\mathbf{x})$ to estimate the risk. We recall, that our main approach involves first computing $N^2$ state-trajectory rollouts and then choosing $N$ out of those for evaluating $h(\mathbf{x})$. This in turn limits the number of collision-checks that we need to perform. For CVaR baselines, the down-sampling process is not relevant and we compute only $N$ state-trajectory rollouts and evaluate $h(\mathbf{x})$ over them. Along similar lines, we also need only $N$ rollouts for the variant of our approach that uses diagonal estimation of RKHS embedding \eqref{empirical_emb_diagonal}. It is worth re-iterating that our approach and all baselines have access to the same $N$ number of samples of constraint $h(\mathbf{x})$ evaluations.
    \end{remark}
    
\vspace{-0.305cm}
\subsection{Benchmarking in Trajectory Optimization Setting}
\noindent We created two benchmark sets with static and dynamic obstacles. For static obstacles, we generated 200 random scenarios and applied two noise models to perturb nominal dynamics. For dynamic obstacles, a single obstacle followed a predefined trajectory, again with two noise models. Qualitative trajectories from these scenarios are shown in the accompanying video. Using Alg.\ref{algo_1}, we computed minimum $r_{MMD}^{emp}$, $r_{MMD-D}^{emp}$ (see \eqref{empirical_emb_diagonal}), and $r_{CVaR}^{emp}$ trajectories for both benchmarks. Crucially, each optimal trajectory satisfied $r_{MMD}^{emp} = 0$, $r_{MMD-D}^{emp} = 0$, and $r_{CVaR}^{emp} = 0$. We then sampled around these optimal trajectories to compute the ground-truth collision rate. To simplify the analysis, this benchmarking did not impose any risk cost for lane violations.


\subsubsection*{\textbf{MMD-vs-CVaR}} The results in Tables \ref{tab:synthetic_static}-\ref{tab:synthetic_dynamic} for different $N$ show that while $r_{MMD}^{emp}$, $r_{MMD-D}^{emp}$, and $r_{CVaR}^{emp}$ are all zero, actual collision rates vary significantly. Trajectories optimized using $r_{MMD}^{emp}$ or $r_{MMD-D}^{emp}$ consistently outperform those from $r_{CVaR}^{emp}$ across all $N$. For instance, in Table \ref{tab:synthetic_static} (static obstacles, high-beta noise, $N=6$), $r_{MMD}^{emp}$ results in a worst-case collision rate that is twice as low as $r_{CVaR}^{emp}$, while $r_{MMD-D}^{emp}$ provides a smaller improvement. A similar trend appears in Table \ref{tab:synthetic_dynamic} for dynamic obstacles—under low-beta noise and $N=6$, MMD-based approaches achieve a median collision rate two to three times lower than the CVaR baseline.

The above result showcases the effectiveness of finite-sample estimate of our risk cost based on MMD vis-a-vis CVaR. This superior performance of $r_{MMD}^{emp}$ can be attributed to the fact the RKHS embedding is particularly effective in capturing the true distribution of arbitrary function of random variables with only a handful of samples (see Fig.1 in \cite{simon2016consistent}). Our optimal \textit{reduced-set} method \eqref{bi_level_red_outer}-\eqref{bi_level_inner_red_cost} further supercharges the capability of the RKHS embedding. 

\subsubsection*{\textbf{$r_{MMD}^{emp}$ vs $r_{MMD-D}^{emp}$}} Table \ref{tab:synthetic_static}-\ref{tab:synthetic_dynamic} shows that $r_{MMD-D}^{emp}$ which is based on diagonal estimation of RKHS embedding (recall \eqref{empirical_emb_diagonal}) can provide similar performance as our main approach based on $r_{MMD}^{emp}$. This is particularly true for low sample regime of $N = 2$, as in this case, the \textit{reduced-set} distillation will not add much value\footnote{For $N=2$, we start with $N^2=4$ samples and down-sample to $N=2$; a difference of just two samples.}. The performance difference is however more important at $N=6$, where $r_{MMD}^{emp}$ clearly shows improved performance.




\subsection{Benchmarking in MPC Setting using CARLA} 
\noindent We now evaluate the efficacy of $r_{MMD}^{emp}$ in a MPC setting where constant re-planning is done based on the current feedback of ego and the neighboring vehicle state. In this benchmarking, we enforce a risk on the lane constraints as well. We also introduced Gaussian noise in the initial state. The qualitative results are presented in the accompanying video.


Table \ref{table_results_obs} presents the quantitative benchmarking in the MPC setting, where we also present results obtained with $r_{MMD-D}^{emp}$. We observed that re-planning can counter some of the effects of noise. 
All approaches improved over the stand-alone planning setting from the previous subsection. However, $r_{MMD}^{emp}$ and $r_{MMD-D}^{emp}$ outperformed $r_{CVaR}^{emp}$ in collision rate and achieved higher max and average speeds. Among them, $r_{MMD}^{emp}$ was superior in all benchmarks except one. The lane violations of CVaR baseline was lower. However, this was due to shorter runs caused by collisions in several experiments. Table \ref{table_results_obs} also includes the deterministic baseline, which, being noise-ignorant, was overconfident, moved too fast, and failed in all simulation runs.

Table \ref{table_results_obs_2} repeats the benchmarking for higher noise level. Here, we only present results for our MMD-based approaches and CVaR baseline. At higher noise, the performance of $r_{MMD}^{emp}$ and $r_{MMD-D}^{emp}$
declines slightly (collision rate, speed) compared to Table \ref{table_results_obs}, but the degradation is significantly lower than in $r_{CVaR}^{emp}$-optimized trajectories.


\begin{table}[!t]
\centering
\captionsetup{justification=centering}
\caption{\footnotesize{$N=2$, Rollout horizon 40, 50 experiments. \\
Gaussian noise: $c_{a,1}=c_{\theta,1}=0.3$, Beta noise: $c_{a,1}=c_{\theta,1}=0.01$. \\
$c_{a,2}=0.3, c_{\theta,2}=0.01$, Gaussian noise in the initial state.}}
\vspace{2mm}
\renewcommand{\arraystretch}{1.5} 
\resizebox{\columnwidth}{!}{%
    \begin{tabular}{|>{\centering\arraybackslash}c|>{\centering\arraybackslash}c|
                    >{\centering\arraybackslash}c|>{\centering\arraybackslash}c|
                    >{\centering\arraybackslash}c|>{\centering\arraybackslash}c|
                    >{\centering\arraybackslash}c|>{\centering\arraybackslash}c|
                    >{\centering\arraybackslash}c|>{\centering\arraybackslash}c|}
    \hline
    \multirow{3}{*}{\textbf{Method}} & \multirow{3}{*}{\textbf{Town}} & \multicolumn{2}{c|}{\textbf{\% Collisions}} & \multicolumn{2}{c|}{\textbf{\% Lane Constr. Viol.}} & \multicolumn{2}{c|}{\textbf{Avg. Speed (m/s)}} & \multicolumn{2}{c|}{\textbf{Max. Speed (m/s)}} \\  
    \cline{3-10}
    & & \textbf{Gaussian} & \textbf{Beta} & \textbf{Gaussian} & \textbf{Beta} & \textbf{Gaussian} & \textbf{Beta} & \textbf{Gaussian} & \textbf{Beta} \\  
    \hline
    $r_{MMD}^{emp}$ & T5 & \textbf{0} & \textbf{0} & 2.39 & 1.18 & 2.59 & 2.68 & 3.69 & 5.08 \\ 
    \hline
    $r_{MMD-D}^{emp}$ & T5 & \textbf{0} & 2.63 & \textbf{0} & 1.23 & 2.05 & 2.72 & 3.26 & 4.79 \\ 
    \hline
    $r_{CVaR}^{emp}$ & T5 & 7.69 & \textbf{0} & 2.25 & 1.08 & 2.12 & 2.49 & 3.04 & 4.61 \\ 
    \hline
    DET & T5 & 100 & 100 & 0.03 & \textbf{0} & \textbf{5.25} & \textbf{5.27} & \textbf{8.72} & \textbf{8.15} \\ 
    \specialrule{2pt}{0pt}{0pt}
    $r_{MMD}^{emp}$ & T10 & \textbf{0} & \textbf{0} & 0.47 & 0.99 & 3.42 & 3.63 & 5.94 & 5.5 \\ 
    \hline
    $r_{MMD-D}^{emp}$ & T10 & \textbf{0} & \textbf{0} & 0.58 & 1.3 & 3.61 & 3.91 & 6.99 & 6.44 \\ 
    \hline
    $r_{CVaR}^{emp}$ & T10 & 16.67 & \textbf{0} & \textbf{0.07} & 0.55 & 2.8 & 3.02 & 5.19 & 5.17 \\ 
    \hline
    DET & T10 & 100 & 100 & 1.01 & \textbf{0.48} & \textbf{5.8} & \textbf{5.93} & \textbf{10.5} & \textbf{9.61} \\ 
    \hline
    \end{tabular}
}
\label{table_results_obs}
\vspace{-0.3cm}
\end{table}

\begin{table}[!t]
\centering
\captionsetup{justification=centering}
\caption{\footnotesize{$N=2$, Rollout horizon 40, 50 experiments. \\
Gaussian noise: $c_{a,1}=c_{\theta,1}=0.3$, Beta noise: $c_{a,1}=c_{\theta,1}=0.05$. \\
$c_{a,2}=0.4, c_{\theta,2}=0.01$, Gaussian noise in the initial state.}}
\vspace{2mm}
\renewcommand{\arraystretch}{1.5} 
\resizebox{\columnwidth}{!}{%
    \begin{tabular}{|>{\centering\arraybackslash}c|>{\centering\arraybackslash}c|
                    >{\centering\arraybackslash}c|>{\centering\arraybackslash}c|
                    >{\centering\arraybackslash}c|>{\centering\arraybackslash}c|
                    >{\centering\arraybackslash}c|>{\centering\arraybackslash}c|
                    >{\centering\arraybackslash}c|>{\centering\arraybackslash}c|}
    \hline
    \multirow{3}{*}{\textbf{Method}} & \multirow{3}{*}{\textbf{Town}} & \multicolumn{2}{c|}{\textbf{\% Collisions}} & \multicolumn{2}{c|}{\textbf{\% Lane Constr. Viol.}} & \multicolumn{2}{c|}{\textbf{Avg. Speed (m/s)}} & \multicolumn{2}{c|}{\textbf{Max. Speed (m/s)}} \\  
    \cline{3-10}
    & & \textbf{Gaussian} & \textbf{Beta} & \textbf{Gaussian} & \textbf{Beta} & \textbf{Gaussian} & \textbf{Beta} & \textbf{Gaussian} & \textbf{Beta} \\  
    \hline
    $r_{MMD}^{emp}$ & T5 & \textbf{0} & \textbf{0} & 2.7 & 3.5 & \textbf{2.59} & 2.02 & \textbf{4.06} & \textbf{4.07} \\ 
    \hline
    $r_{MMD-D}^{emp}$ & T5 & 3.28 & 3.33 & \textbf{2.09} & 7.9 & 2.1 & \textbf{2.1} & 3.7 & \textbf{4.07} \\ 
    \hline
    $r_{CVaR}^{emp}$ & T5 & 15 & 45 & 2.14 & \textbf{2.46} & 2.08 & 1.53 & 3.45 & 3.4 \\ 
    \specialrule{2pt}{0pt}{0pt}
    $r_{MMD}^{emp}$ & T10 & 4 & \textbf{0} & 0.5 & 5.11 & \textbf{3.56} & 2 & 7.01 & 3.73 \\ 
    \hline
    $r_{MMD-D}^{emp}$ & T10 & \textbf{2.44} & \textbf{0} & 0.94 & 8.4 & 3.55 & \textbf{2.29} & \textbf{7.17} & \textbf{4.24} \\ 
    \hline
    $r_{CVaR}^{emp}$ & T10 & 17 & 4 & \textbf{0.1} & \textbf{3.23} & 2.73 & 1.56 & 4.55 & 4.07 \\ 
    \hline
    \end{tabular}
}
\label{table_results_obs_2}
\vspace{-0.5cm}
\end{table}

\subsection{Why MMD approaches performed better than CVaR?} 
\noindent CVaR captures tail risk but requires quantile selection and more samples for stability. In contrast, $r_{MMD}^{emp}$ uses RKHS embedding for richer statistical information, ensuring more accurate risk estimates especially in low-data regimes. Its optimal reduced-set selection further enhances sample efficiency, unlike CVaR.


\subsection{Computational Aspects}

\noindent The $r^{emp}_{CVaR}$ baseline had the lowest computation time (0.1s per MPC step), while $r_{MMD}^{emp}$ took 0.18s due to the overhead from \textit{reduced-set} computation. A simple way to reduce this is by running Alg.\ref{algo_1} for fewer iterations, as shown in the accompanying video. Notably, the simplified $r_{MMD-D}^{emp}$ ran as fast as $r^{emp}_{CVaR}$ but performed significantly better, only slightly worse than $r_{MMD}^{emp}$ optimization. This trend highlights how our MMD-based methods balance performance and re-planning flexibility within an MPC setup. A detailed run-time comparison is provided in the video.

\section{Conclusions and Future Work}
Estimating state-dependent risk through forward simulation is a popular approach in motion planning. But computing rollouts can be expensive. More critically, evaluating constraints over the rollouts could be even more computationally prohibitive, for example,  due to expensive collision checks. Thus, we presented a first such principled approach for reducing the number of constraint evaluations needed to estimate risk. Furthermore, our work can produce reliable results with a very few number of rollouts of system dynamics. Specifically, we leveraged the solid mathematical foundations of RKHS embedding to propose a risk surrogate whose finite sample effectiveness can be further improved by computing the so-called \textit{reduced-set}. We performed extensive simulations in both stand-alone trajectory optimization as well as MPC setting with a strong baseline based on CVaR. We showed that our MMD-based approach outperforms it in collision-rate and achieved average and maximum speed. 


Currently the entire computations of Alg.\ref{algo_1} can run between 5-10 Hz on a RTX 3090 desktop. We believe this performance can be improved by learning good warm-start for Alg.\ref{algo_1}. We are also looking to replace optimization \eqref{bi_level_red_outer}-\eqref{bi_level_inner_red_cost} with a neural network that can directly predict the optimal \textit{reduced-set}. Our future works also seek to extend the formulation to different robotics systems like quadrotors, manipulators, etc.

\bibliography{references}
\bibliographystyle{IEEEtran}

\end{document}